\documentclass[conference]{IEEEtran}
\IEEEoverridecommandlockouts
\usepackage{cite}
\usepackage{adjustbox}
\usepackage{amsmath,amssymb,amsfonts}
\usepackage{algorithmic}
\usepackage{graphicx}
\usepackage{textcomp}
\usepackage{xcolor}
\def\BibTeX{{\rm B\kern-.05em{\sc i\kern-.025em b}\kern-.08em
    T\kern-.1667em\lower.7ex\hbox{E}\kern-.125emX}}
\begin{document}
\title{The Explanation Necessity for Healthcare AI}
\author{\IEEEauthorblockN{Michail Mamalakis}
\IEEEauthorblockA{\textit{Department of Psychiatry,}\\
\textit{Department of Computer Science }\\
\textit{and Technology,} \\
\textit{University of Cambridge}\\
\textit{Cambridge, United Kingdom}\\
\textit{mm2703@cam.ac.uk}}
\and
\and
\and
\and
\IEEEauthorblockN{Héloïse de Vareilles}
\IEEEauthorblockA{\textit{Department of Psychiatry,} \\
\textit{University of Cambridge}\\
\textit{Cambridge, United Kingdom}\\
\textit{hd488@cam.ac.uk}
}
\and
\and
\and
\and
\and
\IEEEauthorblockN{Graham K. Murray}
\IEEEauthorblockA{\textit{Department of Psychiatry,} \\
\textit{University of Cambridge}\\
\textit{Cambridge, United Kingdom}\\
\textit{gm285@cam.ac.uk}
}
\and
\and
\and
\and
\and
\IEEEauthorblockN{Pietro Lio}
\IEEEauthorblockA{\textit{Department of Computer Science }\\
\textit{and Technology,} \\
\textit{University of Cambridge}\\
\textit{Cambridge, United Kingdom}\\
\textit{pl219@cam.ac.uk}
}
\and
\and
\and
\and
\and
\and
\and
\and
\and
\IEEEauthorblockN{John Suckling}
\IEEEauthorblockA{\textit{Department of Psychiatry,} \\
\textit{University of Cambridge}\\
\textit{Cambridge, United Kingdom}\\
\textit{js369@cam.ac.uk}
}
}
\maketitle
\begin{abstract}
Explainability is a critical factor in enhancing the trustworthiness and acceptance of artificial intelligence (AI) in healthcare, where decisions directly impact patient outcomes. Despite advancements in AI interpretability, clear guidelines on when and to what extent explanations are required in medical applications remain lacking. We propose a novel categorization system comprising four classes of explanation necessity (self-explainable, semi-explainable, non-explainable, and new-patterns discovery), guiding the required level of explanation; whether local (patient or sample level), global (cohort or dataset level), or both. To support this system, we introduce a mathematical formulation that incorporates three key factors: (i) robustness of the evaluation protocol, (ii) variability of expert observations, and (iii) representation dimensionality of the application. This framework provides a practical tool for researchers to determine the appropriate depth of explainability needed, addressing the critical question: \textit{When does an AI medical application need to be explained, and at what level of detail?}.
\end{abstract}
\section{Introduction}
Explainable artificial intelligence (XAI) has become a critical concern in digital devices and artificial intelligence (AI) affecting various fields like environmental science, climate studies, automotive technology, and medicine. In particular, the use of XAI in medical practice is crucial due to its significant role in the diagnosis of disease and the care patients receive. XAI plays a key role in fostering trust in algorithms as it aids in understanding risks and identifies therapeutic targets. Additionally, it offers insights into disease progression, treatment response, decision-making, and enables closed-loop control. To this end, a robust explanation of an AI framework can contribute to the design of safety parameters for regulatory consideration of potential therapies \cite{manif}.
\par Although many studies have proposed methods to enhance the interpretability of AI systems \cite{xai},\cite{mx1},\cite{mx2},\cite{mine}, there remains a gap regarding when and at what level explanation is truly required. Specifically, the literature lacks practical guidance on distinguishing between when the explanation necessity required is  for predictions of individual patients or samples, the \textit{local} level, and when it is required to decode the entire model for predictions of the whole cohort or dataset, the \textit{global} level \cite{manif}. In this perspective, we address this question of explanation necessity in the field of AI with a focus on medical applications. We present a categorization that identifies different explanation needs across a range of AI tasks, providing clear algorithmic guidance on when to utilize none, local, global, or both types of explanations. We parameterize the classes of explanation necessity based on the robustness of the evaluation protocol, the degree of agreement among experts' observations, and the representational dimensionality of the application. We propose a mathematical representation of the different categories and discuss various frameworks for delivering explanations. Additionally, we explore different AI tasks and provide examples using our framework.
\par This study does not involve any data or code; rather, it presents a perspective on a categorization system designed to guide the level of explanation necessity.
\section{Methodology}
\subsection{Proposed categories of explanation necessity}
This perspective study categorizes explanation necessity into four distinct classes, tailored to the specific needs of the healthcare and medical domain. These categories are defined by key factors: the robustness of evaluation protocols, variability in expert opinions, and the dimensionality of tasks. The framework reflects our perspective, informed by the unique challenges and priorities of medical contexts, as highlighted in \cite{survey_medical_XAI}. It addresses two critical goals: enabling the exploration of new knowledge for research and development purposes, and supporting the validation of model outputs against established clinical standards in both routine pathological cases and high-risk, time-sensitive scenarios. Our proposed categories are:
\begin{enumerate}
\item \textbf{Self-explainable} applications require no explanation due to very low experts' observation variability, a robust evaluation protocol, low representation dimensionality, and direct understanding of predictions.
\item \textbf{Semi-explainable} applications need local explanations to support training, as they involve low experts' observation variability, robust protocols, and low-medium representation dimensionality.
\item \textbf{Non-explainable} applications require both local and global explanations due to high experts' observation variability, weak protocols, and medium-high representation dimensionality.
\item \textbf{New-patterns discovery} applications demand comprehensive explanations, combining local and global methods, to address significant experts' observation variability, weak protocols, high dimensionality, and the need of new patterns.
\end{enumerate}
\subsection{Main parameters of the framework}
\par \textbf{Variability in expert observations:} To assess the variability in experts' observations, we propose capturing the diversity in annotations or answers provided by experts for each case. In this manuscript we adapt the Guidelines for Reporting Reliability and Agreement Studies (GRRAS) terminology. We primarily focus on "Agreement" which denotes the degree to which scores or observations are the same, and "Inter-rater (or inter-observer) agreement" which signifies the degree to which two or more observers achieve similar results under similar assessment conditions \cite{KOT}. 
A common approach for scoring inter-observer agreement is the calculation of Kappa ($\kappa$) statistics and their variations, including Cohen’s, Fleiss’s, Light’s, and weighted $\kappa$, as reported in thirty-one prior studies (0.39) \cite{review}. Additionally, the utilization of the Landis and Koch interpretation of $\kappa$ is prevalent, found in forty-four prior studies (0.56) \cite{review}.
\begin{figure}
\centerline{
    \includegraphics[trim={0.cm 0.0cm 0.0cm 0.0cm},clip,scale=0.25]{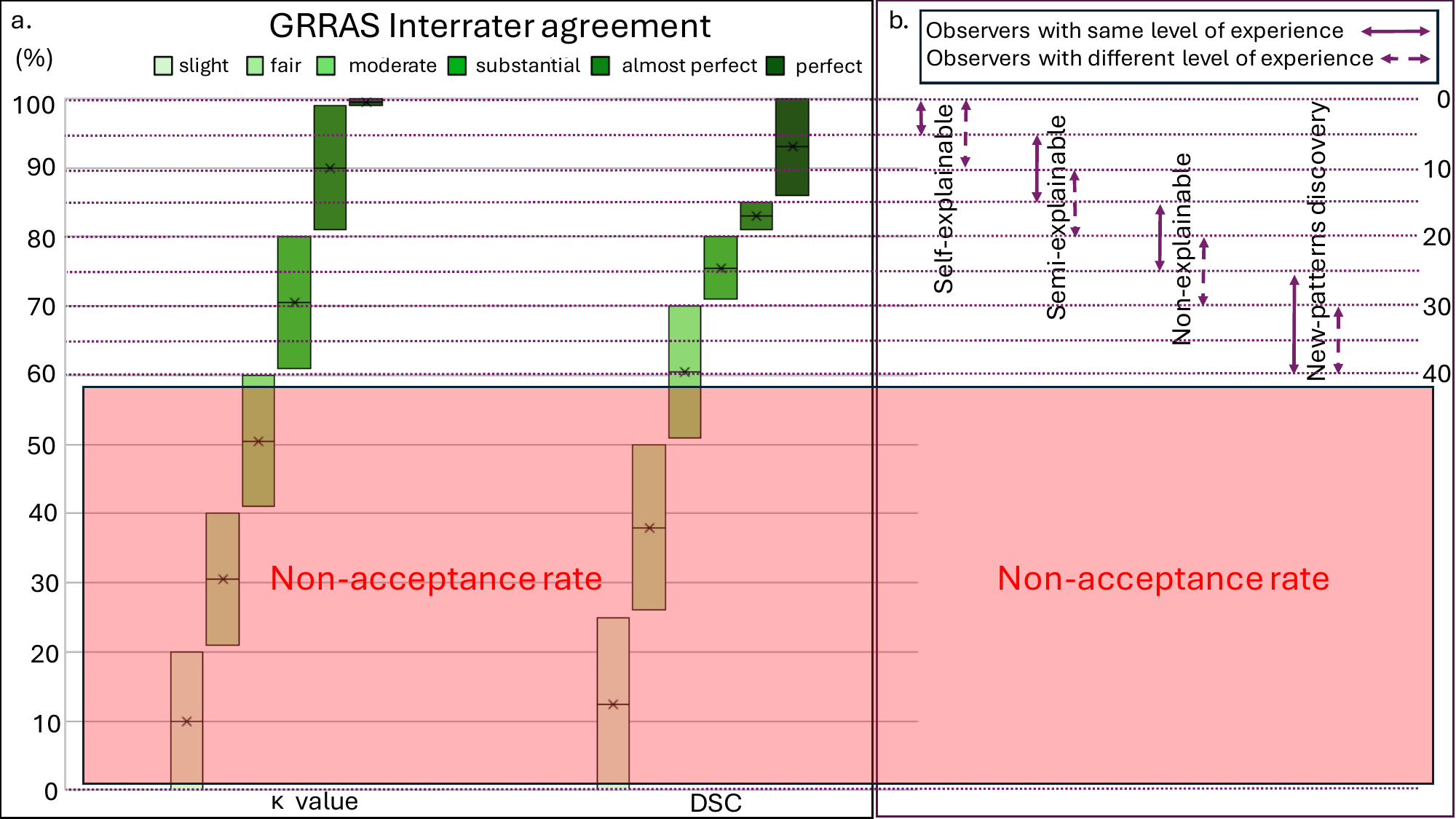}} 
\caption{\textbf{a.} GRRAS $\kappa$-values and Dice Similarity Coefficient (DSC) scores define inter-observer agreement as "slight" to "almost perfect." Low agreement (red box) is unsuitable for medical applications. \textbf{b.} Explanation necessity is categorized based on: (i) variability in expert observations, and (ii) evaluation protocol robustness. Thresholds range from 0.00–0.10 (self-explainable) to 0.31–0.40 (new pattern discovery), varying across and within experience levels (Inexperienced, Experienced, Expert).}
         \label{fc}
   \end{figure}
\par In medical applications, inter-observer variability (observers with same level of experience), a $\kappa$-value between 0.00 and 0.20 is classified as "slight," while values between 0.21 and 0.40 are deemed "fair." "Moderate" agreement falls between 0.41 and 0.60, while "substantial" agreement ranges from 0.61 to 0.80, and "almost perfect" agreement is between 0.81 and 1.00 \cite{KOT, review}. Generally, values of 0.60, 0.70, or 0.80 serve as the minimum standards for the labels for reliability coefficients, but higher values like 0.90 or 0.95 are recommended for critical individual decisions \cite{kott, alma, bbb}. To take as an example the segmentation of lesions or other pathologies from medical images, the agreement DSC is utilized with thresholds: DSC $\geq$ 0.85 considered "High Agreement," 0.85 $>$ DSC $\geq$ 0.70 as "Medium Agreement," 0.7 $>$ DSC $\geq$ 0.5 as "Low Agreement," and DSC $<$ 0.5 as "Very Low Agreement" \cite{tumor, lung} (see Fig. \ref{fc}a.). The proposed number of experts is two to four with the same level of experience in the topic of interest. 
\par \textbf{Robustness of the evaluation protocol:} To assess the robustness of the evaluation protocol we suggest measuring the variability among observers with varying levels of experience (Inexperienced, Experienced, Expert). A robust evaluation protocol is defined by low variability in responses, indicating a clear, well-defined explainable protocol that can be adapted to different experience levels. To this end, we modify the proposed boundaries of the GRRAS Inter-rater agreement discussed above by $\pm$ 5\% to account for the uncertainty arising from varying levels of experience. Typically, a suitable sample size for obtaining robust results consists of two to four observers selected across differing levels of experience (see Fig. \ref{fc}b.). Fig. \ref{fc}b. presents the thresholds that categorize the explanation needs of an AI application based on the robustness of the evaluation protocol (variability in observer experience, 'purple dashed line') and the variability in expert opinions ('purple line'). The thresholds are set according to the level of uncertainty given by the probability $1 - \kappa$ (for classification, regression, etc.) or $1 - DSC$ (for segmentation, registration, overlapping regions etc.) value (see Fig. \ref{fc}b.), that can be tolerated for a specific task, helping to classify the explanation requirements for different AI applications. These thresholds may vary depending on the application, as in some cases diversity of experts' opinion can be significant (like in survival protocols or critical individual decisions \cite{kott, alma, bbb}).
\begin{figure*}
\medskip     
\centerline{
\relax \textbf{a}
    \includegraphics[trim={0.0cm 0.0cm 0.0cm 0.0cm},clip,scale=.32]{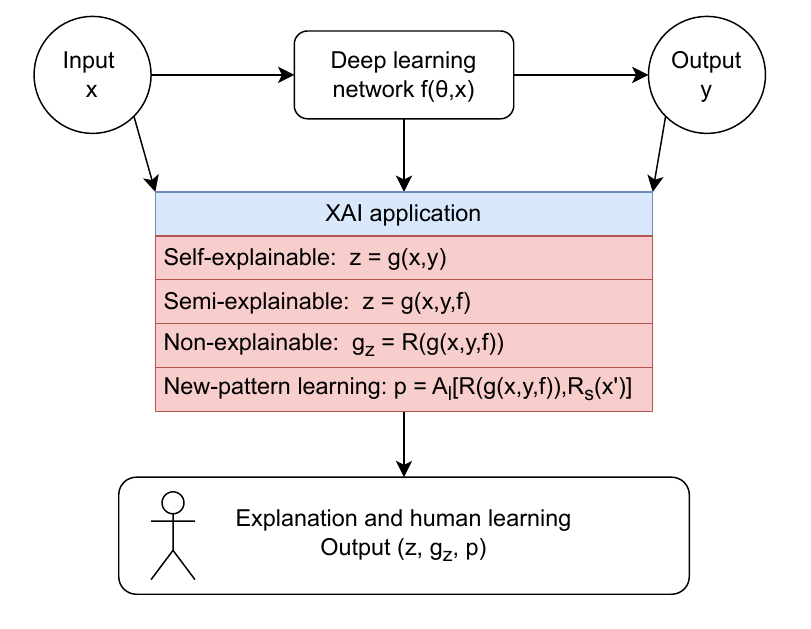}
\relax \textbf{b}
    \includegraphics[trim={0.0cm 0.0cm 0.0cm 0.00cm},clip,scale=.32]{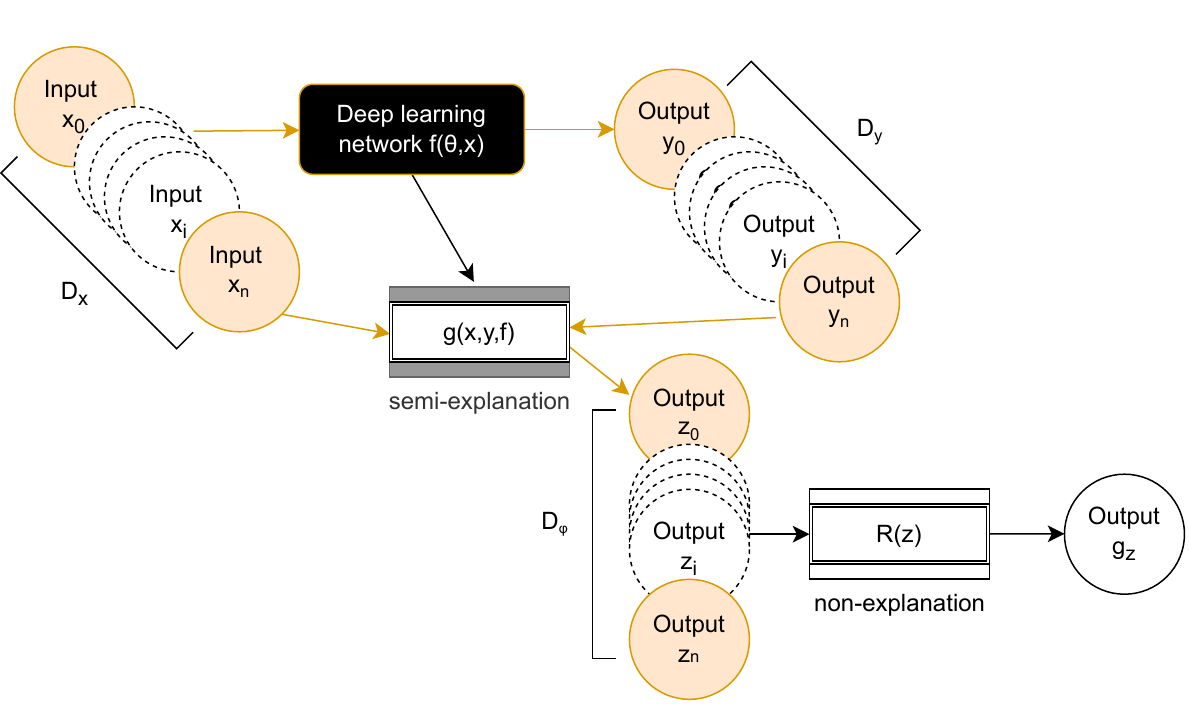}
\relax \textbf{c}
    \includegraphics[trim={0.75cm 0.0cm 0.0cm 0.0cm},clip,scale=.32]{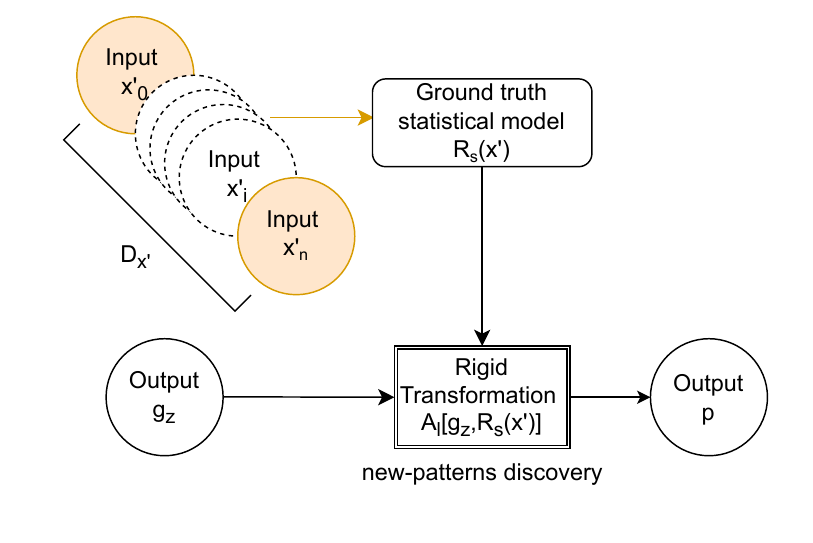}
    }
    \vspace{0.5cm}
    \centerline{
\relax \textbf{d}
    \includegraphics[trim={0.0cm 0.0cm 0.0cm 0.0cm},clip,scale=.3]{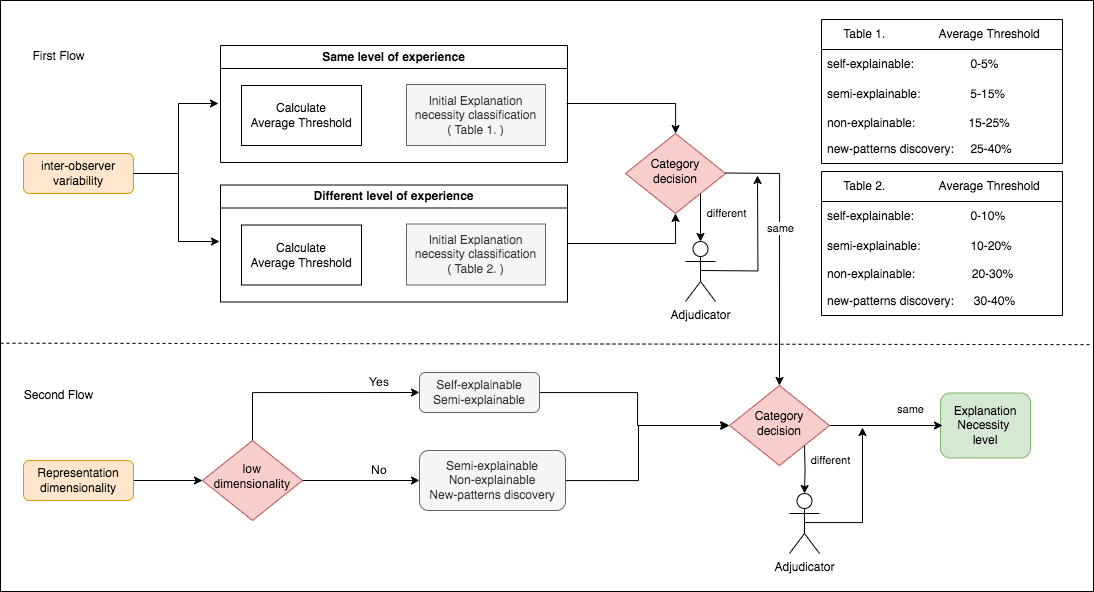}
    }
\small
\caption{
\textbf{a.} The mathematical representation presents the overall XAI framework for a specific method $g$ across different explainability applications.
\textbf{b.} Non-explanations involve methods $g(x,y,f)$ relying on inputs $x$, outputs $y$, and hidden parameters of $f(\theta,x)$. Significant annotation variance and discrepancies in ground truth extraction require global explanations $g_z$ using the entire dataset $D = {(x_i, y_i, z_i)}{i=1}^{n}$.
\textbf{c.} In new-pattern discovery applications, there is the need to align global explanations $g_z$ with a ground truth statistical model $R(x')$ generated from a supergroup $D{x'}$. A transformation function $A_l(g_z,r_{x'})$ aligns the statistical shape model output with the global explanations.
\textbf{d.} Proposed framework for explanation necessity, consisting of two flows: inter-observer variability assessment and dimensionality representation. Variability thresholds (Fig. \ref{fc}) classify 'Initial Explanation Necessity,' with adjudicating experts resolving discrepancies. Dimensionality evaluation refines the final 'Explanation Necessity Level' based on alignment or expert input.}
  \label{math}
\end{figure*}
\par Based on the two parameters of variability in protocols and expert opinions, the proposed categories are defined as presented in Fig. \ref{fc}. Self-explainable tasks have established protocols (0.00-0.10 observer variability) and low expert variability (0.00-0.05; Fig. \ref{fc}b). Semi-explainable tasks have established protocols (0.11-0.20) and low-to-medium expert variability (0.06-0.15; Fig. \ref{fc}b). Non-explainable tasks lack established protocols (0.21-0.30) and have medium-to-high expert variability (0.16-0.25; Fig. \ref{fc}b). New-patterns discovery applies when protocols are unstable (0.31-0.40) and expert variability is high (0.26-0.40; Fig. \ref{fc}b). Thresholds below 0.60 indicate invalid acceptance due to annotation errors or protocol flaws.
\par In studies involving high risk and critical individual decisions \cite{kott, alma, bbb}, it becomes imperative to adapt the proposed thresholds accordingly. In such cases, the acceptable 'Inter-rater agreement' values should ideally surpass the standard thresholds of 0.70, 0.80, or even 0.95, serving as the minimum benchmarks for reliability coefficients (the standard threshold for acceptance rate for 'Inter-rater agreement' typically exceeds 0.60; see Fig. \ref{fc}.). This adjustment ensures a heightened level of reliability and robustness in decision-making processes, crucial for maintaining safety and minimizing potential risks.
\par \textbf{Representational dimensionality of AI applications:} Explanations in AI applications often correlate with the complexity of their representation dimensionality, influencing the level of explanation required. Low-dimensional medical applications, such as structural imaging data (computed tomography, X-rays, magnetic resonance imaging, etc.) or single-modality inputs, fall into semi-explainable and self-explainable tasks involving simpler protocols \cite{mine2, 3dmri}. In contrast, applications with higher complexity, like time-series phenotypic, longitudinal data with a rational structure, or multi-modal inputs, fall into non-explainable or new pattern discovery categories requiring intricate protocols and annotations. These challenges demand evaluation frameworks beyond traditional explanation approaches of only local explanations as such methods may prove insufficient or erroneous in complex domains. For accurate assessment, dimensionality representation must align with explanation needs and the mathematical formulation.
\subsection{Mathematical formulation of the problem}
For an input $\mathbf{x} \in R^d $, we define a deep learning model function $f(\mathbf{x}$; $\mathbf{\theta})$ with $f : R^d \rightarrow R^c$, where $c$ is the output dimension, $d$ is the input dimension and $ \theta $ consists of the parameters of the model (e.g. for a classification task $c$ is the number of classes, and for a computer vision segmentation task $c\leq d$). The inference of the model is denoted as $y = f( \mathbf{x}$; $\mathbf{\theta})$ where $\mathbf{y} \in R^c $ is the predicted probability of the corresponding dimension $c$. An explanation method $g$ from a post-hoc family of explanation methods $G$ takes as inputs the $f$, $\mathbf{x}$ and $\mathbf{y}$ and returns an explanation map $\mathbf{z}$. 
The $G(\mathbf{x},\mathbf{y},f)$ contains multiple explanation methods, denoted as $g^j$ where $j$ is the index of each post-hoc method. The explanation map of each input $\mathbf{x}$ can be given by $z=g(\mathbf{x},\mathbf{y},f)$ where $\mathbf{z} \in R^d$ and has the same dimension space as the input $\mathbf{x}$. We define $D : \mathbb{R}^d \times \mathbb{R}^c \times \mathbb{R}^d \mapsto \mathbb{R}{\geq 0}$ as a tuple of datasets $D = {(\mathbf{x}_i, \mathbf{y}_i, \mathbf{z}_i)}_{i=1}^{n}$, that denotes input-output-explanations triples, where $n$ is the total number of the validation examples. $D_x$ denotes all $\mathbf{x}_i$, $D_y$ denotes all $\mathbf{y}_i$ and $D_z$ denotes all $\mathbf{z}_i$ in $D$. In this study, we define as \textit{"global explanation"} a map $g_z$ that represents all the local explanations for the whole validation dataset $D_z$. The global explanation can be viewed as a dimensionality reduction function $g_z=R(\mathbf{z})$ of the whole local explanations $D_z$ (for example, in a setting of a principal component analysis, $g_z$ it would correspond to the eigenvectors). We define the mathematical formalization of the four proposed categories bellow (Fig. \ref{math} a). 
\par The mathematical representation of the \textbf{self-explainable} application involves an interpretable function $g(\mathbf{x},\mathbf{y})$ that solely employs the inputs $\mathbf{x}$ and outputs $\mathbf{y}$ of the deep learning model $f$. The mathematical description of the \textbf{semi-explainable} application comprises a set of explanation methods $g(\mathbf{x},\mathbf{y},f)$ that utilize both the inputs $\mathbf{x}$ and outputs $\mathbf{y}$  along with the hidden parameters of the deep learning model, $f(\mathbf{x}$; $\mathbf{\theta})$. Due to the  insufficient correlation between the inputs and outputs of the deep learning network, a sub-group $S_D = {(\mathbf{x}_i, \mathbf{y}_i,\mathbf{z}_i)}_{i=1}^{o}$ of local explanations $\mathbf{z}$ of the hidden parameters is needed (where $o$ represents a fixed number of samples from the validation dataset D, $S_D \subset D$ and $o<n$). The mathematical representation of \textbf{non-explainable} applications involves a set of explanation methods $g(\mathbf{x},\mathbf{y},f)$ that relies on the inputs $\mathbf{x}$ and outputs $\mathbf{y}$ in addition to the hidden parameters of the deep learning model $f(\mathbf{x}$; $\mathbf{\theta})$. In this case, the correlation between the inputs and outputs of the deep learning network is inadequate. Thus, there is a requirement for local and additionally global explanations, $g_z$, by using the whole validation dataset $D = {(\mathbf{x}_i, \mathbf{y}_i,\mathbf{z}_i)}_{i=1}^{n}$ (where $n$ encompasses the total number of samples from the validation dataset $D$; Fig. \ref{math} b). In the \textbf{new-patterns discovery} application, a collection of potentially significant markers can be computed for the initial AI task by aligning the global explanations $g_z$ with a ground truth statistical model $R_S(\mathbf{x'})$. This statistical model is generated by inputs $\mathbf{x'}$ of a super dataset $D_{x'}$ of the validation dataset $D_x$ ($D_x \subseteq D_{x'}$). The ground truth statistical model tries to capture the generalized features of interest and the generalized shape related to the AI task. The model's inference is given by $r_{\mathbf{x'}}=R_S(\mathbf{x'})$, where $r_{x'}$ is the output of the ground truth statistical model that best describes the $D_{x'}$ dataset. The alignment can be done by a rigid transformation function $A_l(g_z,r_{\mathbf{x'}})$ of the $r_{\mathbf{x'}}$ and the  $g_z$ (Fig. \ref{math} c). 
\subsection{Proposed framework}
In this study, we propose a framework to classify the explanation necessity of AI medical applications, comprising two main flows: inter-observer variability assessment (First Flow) and representation dimensionality evaluation (Second Flow) (see Fig. \ref{math}d.). The first flow requires users to compute the average inter-observer variability for groups of observers with the same and different experience levels to assess expert variability ('Same level of experience') and the robustness of the evaluation protocol ('Different levels of experience'; Fig. \ref{math}d.). Based on these averages ('Calculate Average Threshold'; Fig. \ref{math}d.), users identify the 'Initial Explanation Necessity Classification' (Fig. \ref{math}d.; Table 1; Table 2). If the classifications differ, an adjudicating expert determines the most appropriate category before proceeding. The second flow evaluates the representation dimensionality of the application ('Representation Dimensionality'). The outcomes of both flows are then passed through an 'XAI Need Decision' statement. If the results align, the final class of explanation necessity is determined ('Category Decision'). If they differ, an adjudicating expert resolves the discrepancy (see Fig. \ref{math}d.). 
\section{Examples and applications}
\par Some medical applications require minimal understanding of AI mechanisms due to low variability in evaluation protocols (0.00-0.10) and expert observations (0.00-0.05). Tasks like organ segmentation from abdominal CT and multi-modal image registration benefit from XAI for optimization rather than trust-building \cite{mine, mine4} (self-explainable applications). In contrast, disease classification tasks (0.05-0.15) require local explanations to ensure proper training and accurate AI assessment \cite{mine2}.
\par In aging populations, neurodegenerative diseases are becoming more prevalent. Binary AI classification from structural MRI for Alzheimer’s diagnosis or healthy aging has low inter-observer variability (0.05-0.15; Fig. \ref{fc}) and low representation dimensionality, making it semi-explainable (Fig. \ref{math}d). Early-stage detection remains significantly more challenging \cite{al}.
\par AI can address knowledge gaps among experts, stabilizing protocols in tasks with unestablished disease evaluation standards (0.25-0.40; Fig. \ref{fc}; \cite{protocol, mine3}). For ovarian cancer, with its uncertain prognosis and challenging early detection even with multi-modal imaging (0.20-0.40; Fig. \ref{fc}; \cite{ova}), AI is categorized as non-explainable or new-patterns discovery. Early sepsis diagnosis, critical for effective treatment, remains similarly complex.
\par Prognosis is particularly poor in areas with limited healthcare access. Chest X-rays \cite{sis1} and whole-body computed tomography \cite{sis2} aid diagnosis and management. However, most AI applications in this domain have high inter-observer variability (0.25-0.40; Fig. \ref{fc}), low evaluation protocol robustness (0.30-0.40; Fig. \ref{fc}), and require multi-modality representation (high dimensionality), classifying them as new-patterns discovery applications.
\section{Conclusion}
Explainability is a critical aspect of AI systems, particularly in healthcare where decisions directly affect patient outcomes.
This study addresses a key gap in the literature by providing user-directed recommendations on when and to what extent explainability techniques should be applied. We propose a novel categorization system comprising four explanation necessity classes: self-explainable, semi-explainable, non-explainable, and new-pattern discovery. These classifications are determined by the variability in expert observations, the robustness of evaluation protocols, and the dimensionality of the application. We proposed a framework incorporates a mathematical formulation to align explanation requirements with application-specific risk levels, adjusting thresholds for high-risk clinical scenarios. By utilizing explanation necessities to application needs, this framework offers practical guidance for researchers and clinicians, ensuring transparency increasing the trustworthiness. It also strengthens safety parameters in regulated medical therapies bridging the gap between theoretical advancements in XAI and their practical implementation in healthcare. The proposed framework applies to various computer vision fields, including natural and automotive vision, but further research is needed to define thresholds for other areas.
\section*{Acknowledgment}
All research at the Department of Psychiatry in the University of Cambridge is supported by the NIHR Cambridge Biomedical Research Centre (NIHR203312) and the NIHR Applied Research Collaboration East of England. The views expressed are those of the author(s) and not necessarily those of the NIHR or the Department of Health and Social Care. This work is supported by the Centre for Human Inspired Artificial Intelligence (CHIA; www.chia.cam.ac.uk). MM is funded by the EU Horizon EBRAINS 2.0 programme.  GKM consults for ieso digital health. All other authors declare that they have no competing interests.
\bibliographystyle{IEEEtran}
\bibliography{IEEEabrv,manuscript}

\begin{thebibliography}{10}
\providecommand{\url}[1]{#1}
\csname url@samestyle\endcsname
\providecommand{\newblock}{\relax}
\providecommand{\bibinfo}[2]{#2}
\providecommand{\BIBentrySTDinterwordspacing}{\spaceskip=0pt\relax}
\providecommand{\BIBentryALTinterwordstretchfactor}{4}
\providecommand{\BIBentryALTinterwordspacing}{\spaceskip=\fontdimen2\font plus
\BIBentryALTinterwordstretchfactor\fontdimen3\font minus
  \fontdimen4\font\relax}
\providecommand{\BIBforeignlanguage}[2]{{%
\expandafter\ifx\csname l@#1\endcsname\relax
\typeout{** WARNING: IEEEtran.bst: No hyphenation pattern has been}%
\typeout{** loaded for the language `#1'. Using the pattern for}%
\typeout{** the default language instead.}%
\else
\language=\csname l@#1\endcsname
\fi
#2}}
\providecommand{\BIBdecl}{\relax}
\BIBdecl

\bibitem{manif}
\BIBentryALTinterwordspacing
L.~Longo, M.~Brcic, F.~Cabitza, J.~Choi, R.~Confalonieri, J.~D. Ser,
  R.~Guidotti, Y.~Hayashi, F.~Herrera, A.~Holzinger, R.~Jiang, H.~Khosravi,
  F.~Lecue, G.~Malgieri, A.~Páez, W.~Samek, J.~Schneider, T.~Speith, and
  S.~Stumpf, ``Explainable artificial intelligence (xai) 2.0: A manifesto of
  open challenges and interdisciplinary research directions,''
  \emph{Information Fusion}, vol. 106, p. 102301, 2024. [Online]. Available:
  \url{https://www.sciencedirect.com/science/article/pii/S1566253524000794}
\BIBentrySTDinterwordspacing

\bibitem{xai}
W.~Samek, G.~Montavon, S.~Lapuschkin, C.~J. Anders, and K.-R. Müller,
  ``Explaining deep neural networks and beyond: A review of methods and
  applications,'' \emph{Proceedings of the IEEE}, vol. 109, no.~3, pp.
  247--278, 2021.

\bibitem{mx1}
\BIBentryALTinterwordspacing
B.~H. van~der Velden, H.~J. Kuijf, K.~G. Gilhuijs, and M.~A. Viergever,
  ``Explainable artificial intelligence (xai) in deep learning-based medical
  image analysis,'' \emph{Medical Image Analysis}, vol.~79, p. 102470, 2022.
  [Online]. Available:
  \url{https://www.sciencedirect.com/science/article/pii/S1361841522001177}
\BIBentrySTDinterwordspacing

\bibitem{mx2}
\BIBentryALTinterwordspacing
G.~Quellec, H.~{Al Hajj}, M.~Lamard, P.-H. Conze, P.~Massin, and B.~Cochener,
  ``Explain: Explanatory artificial intelligence for diabetic retinopathy
  diagnosis,'' \emph{Medical Image Analysis}, vol.~72, p. 102118, 2021.
  [Online]. Available:
  \url{https://www.sciencedirect.com/science/article/pii/S136184152100164X}
\BIBentrySTDinterwordspacing

\bibitem{mine}
\BIBentryALTinterwordspacing
M.~Mamalakis, P.~Garg, T.~Nelson, J.~Lee, J.~M. Wild, and R.~H. Clayton,
  ``Ma-socratis: An automatic pipeline for robust segmentation of the left
  ventricle and scar,'' \emph{Computerized Medical Imaging and Graphics},
  vol.~93, p. 101982, 2021. [Online]. Available:
  \url{https://www.sciencedirect.com/science/article/pii/S0895611121001312}
\BIBentrySTDinterwordspacing

\bibitem{survey_medical_XAI}
\BIBentryALTinterwordspacing
N.~Bienefeld, J.~M. Boss, R.~L{\"u}thy, D.~Brodbeck, J.~Azzati, M.~Blaser,
  J.~Willms, and E.~Keller, ``Solving the explainable ai conundrum by bridging
  clinicians'needs and developers'goals,'' \emph{npj Digital Medicine}, vol.~6,
  no.~1, p.~94, 2023. [Online]. Available:
  \url{https://doi.org/10.1038/s41746-023-00837-4}
\BIBentrySTDinterwordspacing

\bibitem{KOT}
\BIBentryALTinterwordspacing
J.~Kottner, L.~Audige, S.~Brorson, A.~Donner, B.~J. Gajewski, A.~Hróbjartsson,
  C.~Roberts, M.~Shoukri, and D.~L. Streiner, ``Guidelines for reporting
  reliability and agreement studies (grras) were proposed,''
  \emph{International Journal of Nursing Studies}, vol.~48, no.~6, pp.
  661--671, 2011. [Online]. Available:
  \url{https://www.sciencedirect.com/science/article/pii/S0020748911000368}
\BIBentrySTDinterwordspacing

\bibitem{review}
\BIBentryALTinterwordspacing
L.~Quinn, K.~Tryposkiadis, J.~Deeks, H.~C. De~Vet, S.~Mallett, L.~B. Mokkink,
  Y.~Takwoingi, S.~Taylor-Phillips, and A.~Sitch, ``{Interobserver variability
  studies in diagnostic imaging: a methodological systematic review},''
  \emph{British Journal of Radiology}, vol.~96, no. 1148, p. 20220972, 06 2023.
  [Online]. Available: \url{https://doi.org/10.1259/bjr.20220972}
\BIBentrySTDinterwordspacing

\bibitem{kott}
\BIBentryALTinterwordspacing
J.~Kottner and T.~Dassen, ``Interpreting interrater reliability coefficients of
  the braden scale: A discussion paper,'' \emph{International Journal of
  Nursing Studies}, vol.~45, no.~8, pp. 1238--1246, 2008. [Online]. Available:
  \url{https://www.sciencedirect.com/science/article/pii/S002074890700199X}
\BIBentrySTDinterwordspacing

\bibitem{alma}
D.~Polit-O'Hara and C.~T. Beck, \emph{\BIBforeignlanguage{eng}{Nursing research
  : generating and assessing evidence for nursing practice / Denise F. Polit,
  Cheryl Tatano Beck.}}, eighth edition.~ed., Philadelphia, 2008.

\bibitem{bbb}
\BIBentryALTinterwordspacing
R.~M. Thorndike, ``Book review : Psychometric theory (3rd ed.) by jum nunnally
  and ira bernstein new york: Mcgraw-hill, 1994, xxiv + 752 pp,'' \emph{Applied
  Psychological Measurement}, vol.~19, no.~3, pp. 303--305, 1995. [Online].
  Available: \url{https://doi.org/10.1177/014662169501900308}
\BIBentrySTDinterwordspacing

\bibitem{tumor}
\BIBentryALTinterwordspacing
J.~Wong, M.~Baine, S.~Wisnoskie, N.~Bennion, D.~Zheng, L.~Yu, V.~Dalal, M.~A.
  Hollingsworth, C.~Lin, and D.~Zheng, ``Effects of interobserver and
  interdisciplinary segmentation variabilities on ct-based radiomics for
  pancreatic cancer,'' \emph{Scientific Reports}, vol.~11, no.~1, p. 16328,
  2021. [Online]. Available: \url{https://doi.org/10.1038/s41598-021-95152-x}
\BIBentrySTDinterwordspacing

\bibitem{lung}
\BIBentryALTinterwordspacing
G.~Kothari, B.~Woon, C.~J. Patrick, J.~Korte, L.~Wee, G.~G. Hanna, T.~Kron,
  N.~Hardcastle, and S.~Siva, ``The impact of inter-observer variation in
  delineation on robustness of radiomics features in non-small cell lung
  cancer,'' \emph{Scientific Reports}, vol.~12, no.~1, p. 12822, 2022.
  [Online]. Available: \url{https://doi.org/10.1038/s41598-022-16520-9}
\BIBentrySTDinterwordspacing

\bibitem{mine2}
\BIBentryALTinterwordspacing
M.~Mamalakis, S.~C. Macfarlane, S.~V. Notley, A.~K. Gad, and G.~Panoutsos, ``A
  novel pipeline employing deep multi-attention channels network for the
  autonomous detection of metastasizing cells through fluorescence
  microscopy,'' \emph{Computers in Biology and Medicine}, vol. 181, p. 109052,
  2024. [Online]. Available:
  \url{https://www.sciencedirect.com/science/article/pii/S0010482524011375}
\BIBentrySTDinterwordspacing

\bibitem{3dmri}
\BIBentryALTinterwordspacing
J.~Ma, Y.~He, F.~Li, L.~Han, C.~You, and B.~Wang, ``Segment anything in medical
  images,'' \emph{Nature Communications}, vol.~15, no.~1, p. 654, 2024.
  [Online]. Available: \url{https://doi.org/10.1038/s41467-024-44824-z}
\BIBentrySTDinterwordspacing

\bibitem{mine4}
\BIBentryALTinterwordspacing
M.~Mamalakis, P.~Garg, T.~Nelson, J.~Lee, A.~J. Swift, J.~M. Wild, and R.~H.
  Clayton, ``Artificial intelligence framework with traditional computer vision
  and deep learning approaches for optimal automatic segmentation of left
  ventricle with scar,'' \emph{Artificial Intelligence in Medicine}, vol. 143,
  p. 102610, 2023. [Online]. Available:
  \url{https://www.sciencedirect.com/science/article/pii/S0933365723001240}
\BIBentrySTDinterwordspacing

\bibitem{al}
\BIBentryALTinterwordspacing
E.~E. Bron, S.~Klein, A.~Reinke, J.~M. Papma, L.~Maier-Hein, D.~C. Alexander,
  and N.~P. Oxtoby, ``Ten years of image analysis and machine learning
  competitions in dementia,'' \emph{NeuroImage}, vol. 253, p. 119083, 2022.
  [Online]. Available:
  \url{https://www.sciencedirect.com/science/article/pii/S1053811922002129}
\BIBentrySTDinterwordspacing

\bibitem{protocol}
\BIBentryALTinterwordspacing
S.~C. Mitchell, H.~De~Vareilles, J.~R. Garrison, A.~Al-Manea, J.~Suckling,
  G.~K. Murray, and J.~S. Simons, ``Paracingulate sulcus measurement protocol
  v2,'' 2023. [Online]. Available:
  \url{https://www.repository.cam.ac.uk/handle/1810/358381}
\BIBentrySTDinterwordspacing

\bibitem{mine3}
\BIBentryALTinterwordspacing
M.~Mamalakis, K.~Dwivedi, M.~Sharkey, S.~Alabed, D.~Kiely, and A.~J. Swift, ``A
  transparent artificial intelligence framework to assess lung disease in
  pulmonary hypertension,'' \emph{Scientific Reports}, vol.~13, no.~1, p. 3812,
  2023. [Online]. Available: \url{https://doi.org/10.1038/s41598-023-30503-4}
\BIBentrySTDinterwordspacing

\bibitem{ova}
\BIBentryALTinterwordspacing
K.~B. Mathieu, D.~G. Bedi, S.~L. Thrower, A.~Qayyum, and R.~C. Bast~Jr,
  ``Screening for ovarian cancer: imaging challenges and opportunities for
  improvement,'' \emph{Ultrasound in Obstetrics \& Gynecology}, vol.~51, no.~3,
  pp. 293--303, 2018. [Online]. Available:
  \url{https://obgyn.onlinelibrary.wiley.com/doi/abs/10.1002/uog.17557}
\BIBentrySTDinterwordspacing

\bibitem{sis1}
\BIBentryALTinterwordspacing
L.~Arias-Alvarez, G.~Ortiz-Ruiz, and C.~Dueñas-Castell, ``Diagnostic imaging
  in sepsis of pulmonary origin,'' \emph{Infectious Disease Clinics of North
  America}, vol.~32, no.~1, pp. 149--165, 2018. [Online]. Available:
  \url{https://www.ncbi.nlm.nih.gov/pmc/articles/PMC7120880/}
\BIBentrySTDinterwordspacing

\bibitem{sis2}
\BIBentryALTinterwordspacing
J.~Pohlan, D.~Witham, M.~I. Opper~Hernando, G.~Muench, M.~Anhamm, A.~Schnorr,
  L.~Farkic, K.~Breiling, R.~Ahlborn, K.~Rubarth, D.~Praeger, and M.~Dewey,
  ``Relevance of ct for the detection of septic foci: diagnostic performance in
  a retrospective cohort of medical intensive care patients,'' \emph{Clinical
  Radiology}, vol.~77, no.~3, pp. 203--209, 2022. [Online]. Available:
  \url{https://pubmed.ncbi.nlm.nih.gov/34872706/}
\BIBentrySTDinterwordspacing

\end{thebibliography}
\end{document}